\pdfoutput=1
\documentclass[11pt]{article}

\usepackage[]{acl}

\usepackage{times}
\usepackage{latexsym}

\usepackage{color}
\usepackage{graphicx}
\usepackage{amsmath}
\usepackage{amssymb}
\usepackage{booktabs}
\usepackage{multirow}
\usepackage{algorithm}
\usepackage{algorithmic}

\usepackage[T1]{fontenc}
\usepackage[utf8]{inputenc}

\usepackage{microtype}

\title{Explainable Topic-Enhanced Argument Mining from Heterogeneous Sources}

\author{Jiasheng Si$^{\dag}$ \ \ \ \  Yingjie Zhu$^{\dag}$ \ \ \ \ Xingyu Shi$^{\dag}$ \ \ \ \ Deyu Zhou$^{\dag}\thanks{corresponding author}$ \ \ \ \   Yulan He$^{\S\ddag}$ \\
$^{\dag}$ School of Computer Science and Engineering, Key Laboratory of Computer Network\\
	and Information Integration, Ministry of Education, Southeast University, China \\
$\S$Department of informatics, King’s College London, UK \\
 $^\ddag$Alan Turing Insitute, UK \\
\texttt{\{jasenchn, yj\_zhu, xyu-shi, d.zhou\}@seu.edu.cn}, \\
\texttt{yulan.he@kcl.ac.uk}}

\begin{document}
\maketitle
\begin{abstract}
	Given a controversial target such as ``\textit{nuclear energy}'',
	argument mining aims to identify
	the argumentative text from heterogeneous sources. Current approaches focus on exploring better
	ways of
	integrating the target-associated semantic information
	with the argumentative text.
	Despite their empirical successes,
	two issues remain unsolved: (i) a target is represented by a word or a phrase, which is insufficient to cover a diverse set of target-related subtopics;
	(ii) the sentence-level topic information within an argument,
	which we believe is crucial for argument mining, is ignored.
	To tackle the above issues, we propose a novel explainable topic-enhanced argument mining approach. Specifically,  with the use of the neural topic model and the language model, the target information is augmented by  explainable topic representations. Moreover, the sentence-level topic information within the argument is captured by minimizing the distance between its latent topic distribution and its semantic representation through mutual learning. %Comprehensive 
	Experiments have been conducted on the benchmark dataset in both the in-target setting and the cross-target setting. Results demonstrate the superiority of the proposed model against the state-of-the-art baselines.

\end{abstract}

\section{Introduction}
Argument mining (AM), aiming to extract and identify argumentative structures from natural language text,
%find applications in areas such as %provides the perfect ground to combine machine learning with decision making, legal 
%reasoning in legal text documents and human debating. It 
has became an established field in the natural language processing community~\citep{DBLP:conf/ijcai/CabrioV18, DBLP:conf/aaai/LippiT16, DBLP:conf/aaai/NguyenL18, DBLP:journals/coling/LawrenceR19, DBLP:conf/acl/VecchiFJL20}.
Broadly speaking, two basic paradigms of argument mining have been proposed: \textit{discourse-level} AM and \textit{information-seeking} AM.
The former mainly focuses on identifying the explicit argumentative structure within a document via~\textit{claim-premise} model~\citep{stab-gurevych-2014-annotating, DBLP:conf/acl/EgerDG17,DBLP:journals/coling/HabernalG17, DBLP:conf/acl/KuribayashiOIRM19}.
Instead of focusing on a single document,
the latter aims to identify self-contained argumentative texts with respect to the controversial targets\footnote{In the following, we use the term ``\emph{target}'' rather than ``\emph{topic}'' usually used in the previous researches to differentiate it from the~\textit{topic} extracted by the topic models.} from heterogeneous sources~\citep{DBLP:conf/argmining/WachsmuthPKAPQD17, stab-etal-2018-cross, DBLP:conf/aaai/TrautmannDSSG20, DBLP:conf/aaai/Ein-DorSDHSGAGC20, slonim2021autonomous},
which can be considered as a remedy to the discourse-level AM.
In this paper, we illustrate how the topic information can be leveraged to deal with the \emph{information-seeking} AM task.
% focus on the \emph{information-seeking} AM and show how the topic information can be leveraged to overcome some of the challenges contained in this task.

\begin{figure}[]
	\centering
	\includegraphics[scale=0.55]{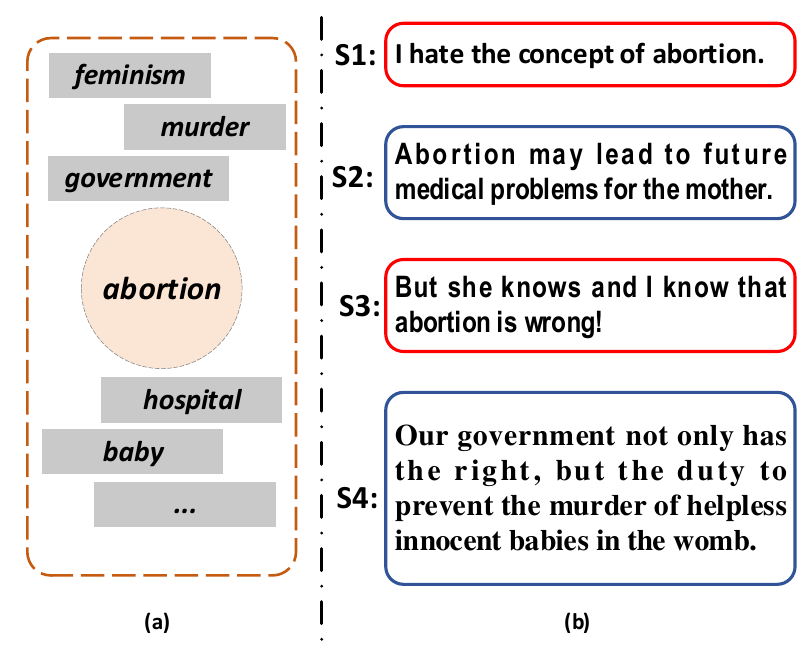}
	\caption{Problem illustration. (a) the given target (i.e., \textit{abortion}) and the extracted topic words from the \textit{UKP ArgMin}. (b) a list of arguments. Here, $S1, S3$ are \textit{none} arguments (express \textit{opinion} with no \textit{reason} provided); $S2, S4$ are \textit{oppose} arguments (express \textit{opinion} and \textit{reason} simultaneously).}
	\label{motivation}
\end{figure}

There are strands of research dealing with the information-seeking AM by integrating the target information with an argument for learning the target-dependent representation of the argument~\citep{stab-etal-2018-cross,DBLP:conf/acl/ReimersSBDSG19, lin-etal-2019-lexicon, DBLP:conf/aaai/TrautmannDSSG20}. One example is to incorporate %introducing 
the semantic representation of the target through the attention mechanism or the recurrent units~\citep{stab-etal-2018-cross}, which is further augmented %augmenting the target 
with synonyms from an external lexicon~\citep{lin-etal-2019-lexicon}.
Moreover,
different from the previous content-only approaches,
\citet{DBLP:conf/acl/BarrowJLDMMORW20} introduced a syntopical graph to capture the collection-level information of arguments.

However, despite the salient progress, the aforementioned methods suffer from two limitations: (1) a target is usually represented by %consists of 
only a word or a phrase %one or two words 
(e.g.,~\textit{abortion, nuclear energy}). Such a representation cannot capture a wide range of information relating to the target. %limits the ability of capturing the semantic information of the target.
Different with the most previous methods focusing on the synonyms to augment the target, we argue that people prefer to discuss~\textit{subtopics} relevant to the target rather than directly using target-related synonyms. %semantic-relevant words. 
As shown in Figure~\ref{motivation},  for the target \textit{abortion}, words such as \textit{feminism, murder, government} are more often mentioned compared to \textit{abandonment, cancellation}, and \textit{failure}. %We believe that 
Therefore, expressions in subtopics can be used to improve the representation of an argument, which also provides a certain level of explanation; %fine-granularity (explanatory) outlook. 
(2) the sentence-level topic information in arguments is usually ignored.
We argue that it might play a crucial role for learning the target-dependent representation of an argument.
As shown in Figure~\ref{motivation}, for the \textit{support} or \textit{oppose} arguments (e.g., \textit{S2, S4}), %more target-related 
related topics discussed in sentences (e.g., \textit{medical, mother} for $S2$ and \textit{government, baby, murder} for $S4$) reveal %illustrate 
the reasons behind the expressed stances, as opposed to %for holding such stances compared to 
the \textit{none} argumentative sentences such as \textit{S1} and \textit{S3}, in which no reason is given.
Therefore, modeling the sentence-level topic information within an argument could improve the performance of information-seeking argument mining.

Motivated by the observations above, we propose a \textbf{T}opic-\textbf{E}nhanced \textbf{A}rgument \textbf{M}ining (TEAM) model for information-seeking AM.
Specifically, with the use of the neural topic model (NTM) and the pre-trained language model (LM),
(1) the target is augmented with the explainable topics (a collection of key terms) which are extracted from the global topic-word distribution; % with the topic filtering and the target-relevant topic extracting;
(2) the sentence-level topic information of the argument is captured by minimizing the distance between its latent topic distribution and its semantic representation through mutual learning.

To sum up, the main contributions of the paper are:
\begin{itemize}
	\item A novel \textbf{T}opic-\textbf{E}nhanced \textbf{A}rgument \textbf{M}ining (TEAM) model is proposed, which is, to our best knowledge, the first attempt of exploring the topic information for information-seeking AM.  The explainable topics are also extracted to show the rationale behind the argument for each target.
	\item Comprehensive experiments are conducted on the \textit{UKP ArgMin} in both the in-target setting and the cross-target setting to show the effectiveness of the proposed model.
\end{itemize}

\section{Related Work}

Previous approaches on argument mining in natural language processing mainly focused on the~\textit{discourse-level} AM~\citep{DBLP:journals/coling/LawrenceR19},
which aims to automatically segment and  identify the argument unit~\citep{stab-gurevych-2014-annotating, DBLP:conf/setn/GoudasLPK14, DBLP:conf/argmining/AjjourCKWS17}
and recognize the relations between the argument pair (i.e.,~\textit{claim-premise})~\citep{DBLP:journals/coling/StabG17, DBLP:conf/acl/NguyenL16, DBLP:conf/acl/EgerDG17}.
However, such approaches make a strong assumption that the explicitly argumentative structure is expressed in one single document.
It is doubtful whether they can be reliably applied to the \textit{open-domain} argument retrieval with different text types and targets.

As a remedy to this, \textit{information-seeking} AM aims to identify the relevant argumentative text towards the controversial target from heterogeneous sources. It has facilitated the development of the argument search engine~\citep{DBLP:conf/acl/Bar-HaimKTEBHKM19, DBLP:conf/sigir/PotthastGEHWWSH19, DBLP:journals/dbsk/DaxenbergerSSKG20, slonim2021autonomous}.
Existing approaches for this task mainly focused on exploring the semantic relevance between the target and the argument.
\citet{levy-etal-2014-context} investigated the identification of topic-relevant claims by splitting the main task into several subtasks.
\citet{rinott-etal-2015-show} extended it with evidence extraction to mine supporting statements from Wikipedia for claims.
\citet{wachsmuth-etal-2017-building} presented a generic argument search framework that relies on already-structured arguments from debate portals.
Instead of incorporating the structure information,
\citet{DBLP:conf/acl/ShnarchPDGHCAS18} identified the target-dependent evidence sentence by blending the generated training sets.
\citet{stab-etal-2018-cross} used the attention mechanism based vanilla LSTM model to fuse the target and candidate argument. Furthermore,
\citet{DBLP:conf/acl/ReimersSBDSG19} identified the target-dependent arguments by leveraging the contextualized word embedding.
\citet{lin-etal-2019-lexicon} integrated external lexicon information from several resources into the representation learning of the argument.
In addition,
\citet{DBLP:conf/aaai/TrautmannDSSG20} introduced the argument identification into the fine-grained argument segmentation.

Different from the existing approaches which incorporate the semantic information of the target into the argument representation learning,
our work for the first time handles information-seeking argument mining from the topical perspective,
where the target is augmented with explainable topic representations and the sentence-level topic information within the argument is precisely captured using the topic-argument mutual learning mechanism.

\begin{figure*}[ht]

	\centering
	\includegraphics[scale=0.6]{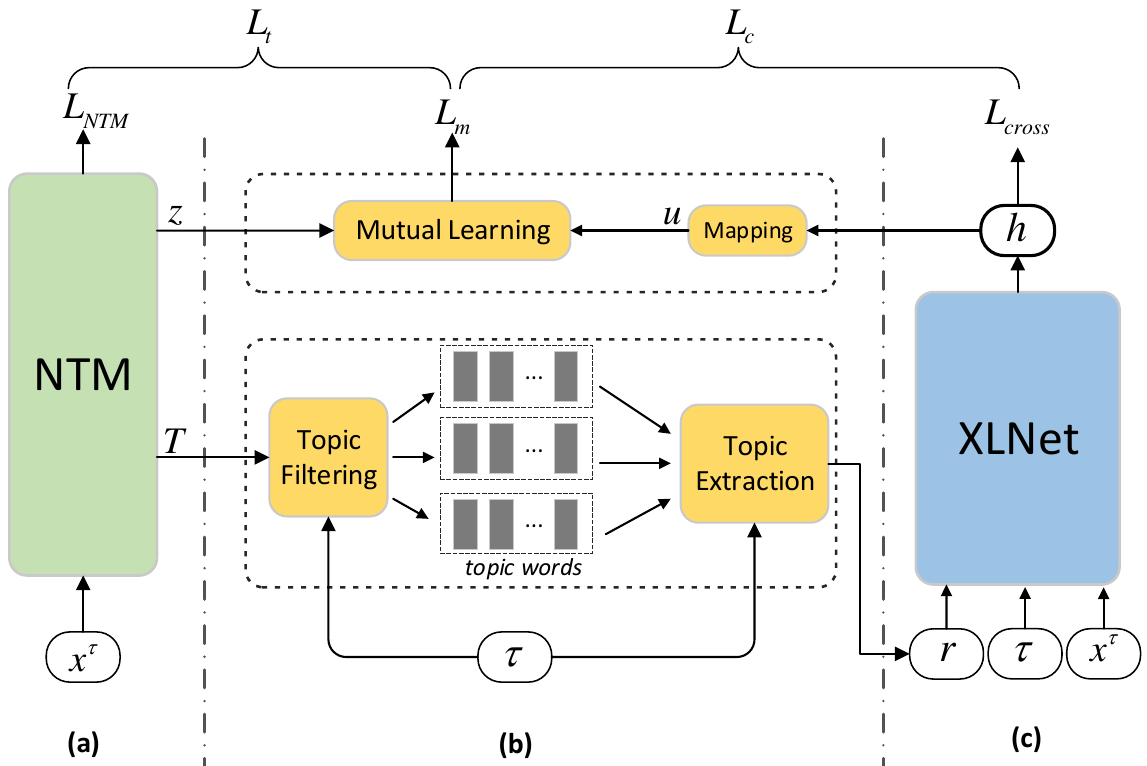}
	\caption{The architecture of the proposed TEAM model: (a) topic representation generation, (b) topic-argument mutual learning (top), explainable topic extraction (bottom), (c) argument identification. }
	\label{fig:overall-structure}
\end{figure*}

\section{Methodology}

An argument is defined as the combination of a target $\tau$ with a sentence $\boldsymbol{x}^{\tau}$.
Based on the \textit{opinion} and \textit{reason} contained in the argument, we aim to develop a model to assign the correct label $y^{\tau}\in\{support, oppose, none\}$ to the argument in two settings. In the in-target setting, the traditional supervised paradigm is employed. In the cross-target setting, one target is withheld from the training data and used for testing.

The overall structure of the proposed model is shown in Figure \ref{fig:overall-structure}, which consists of four main components:
1) topic representation generation, which infers the latent topic distribution and the global topic-word distribution using a neural topic model (Figure~\ref{fig:overall-structure}(a));
2) explainable topic extraction, which extracts the explainable topics from the global topic-word distribution (bottom
of Figure~\ref{fig:overall-structure}(b));
3) argument identification, which learns the semantic representation of the argument (Figure~\ref{fig:overall-structure}(c));
4) topic-argument mutual learning, which captures the sentence-level topic information by minimizing the distance between the latent topic distribution and the semantic representation within an argument via mutual learning (top of Figure~\ref{fig:overall-structure}(b)).
In what follows, we will describe each of the components in turn.

\subsection{Topic Representation Generation}

To exploit the use of topic knowledge for argument identification, %while identifying argument,
we first extract topics using NTM~\citep{DBLP:conf/icml/MiaoGB17, DBLP:conf/iclr/SrivastavaS17}. NTM is a latent Dirichlet allocation (LDA) based topic model built on the %via 
variational autoencoder (VAE)~\cite{DBLP:journals/corr/KingmaW13}.

Let $\boldsymbol{v}\in\mathbb{R}^V$ denotes the bag-of-words representation of an argument sentence $\boldsymbol{x}^{\tau}\in D$,
where $V$ denotes the vocabulary size.

\paragraph{Inference Network}
Following the idea of VAE, we define two multi-layer perceptrons (MLP) which take $\boldsymbol{v}$ as input and output the mean vector $\mu_{\theta} (\boldsymbol{v})={\rm MLP}_\mu(\boldsymbol{v})$ and variance vector $\sum_{\theta}  (\boldsymbol{v})={\rm MLP}_\sigma(\boldsymbol{v})$ of a Gaussian distribution. The latent variable $\hat{\theta}$ is generated using the re-parameterization trick, $\hat{\theta}=\sigma(\mu_{\theta}+\sum_{\theta}^{1/2}\epsilon)$,
where $\epsilon\sim N(0, I^2)$.

\paragraph{Generation Network}
We feed the latent variable $\hat{\theta}$ to the \textit{MLP} to generate the latent topic distribution $\boldsymbol{z}\in\mathbb{R}^K$,
where $K$ is the number of topics.
Then the probability of drawing the word $w_n$ is computed by,
\begin{align}
	p(w_n|\boldsymbol{v},\boldsymbol{z})\propto exp(\boldsymbol{m} + \boldsymbol{T}\cdot \boldsymbol{z})
\end{align}
where $\boldsymbol{m} \in \mathbb{R}^V$ denotes the log-frequency word distribution, $\boldsymbol{T}\in \mathbb{R}^{V\times K}$.

During training, the VAE-based topic model aims to maximize %estimate 
the Evidence Lower Bound with a Monte Carlo approximation:
\begin{align}
	 & \mathcal{L}_{\mathrm{NTM}}\approx \\ &
	\frac{1}{L}\sum^{L}_{l=1}\sum_{n=1}^{N}\log{p(w|\hat{\theta}^{(l)})} - KL(q(\boldsymbol{z}|\boldsymbol{w})||p(\boldsymbol{z}))\notag
\end{align}
where $L$ denotes the number of sentences and $N$ denotes the number of words within a sentence.

Finally, the sentence-level latent topic distribution $\boldsymbol{z}\in \mathbb{R}^K$ and the global topic-word distribution $\boldsymbol{T}\in \mathbb{R}^{V\times K}$ are extracted as the output of this component.

\subsection{Explainable Topic Extraction}
\label{sec: target-specific}

As the controversial target consists of either a single word or a multi-word phrase, %one or two words, 
which is insufficient to capture the information of the target,
inspired by~\citep{DBLP:conf/icml/ChaudharySG20}, we extract the explainable topics to enhance the representation of the target,
which is different with the previous methods focusing on the synonyms of target.
In specific, the explainable topics are sampled from a set of high-probability terms in its topic-word distribution.
The rationale behind this is that the~\textit{subtopics} relevant to the target are often mentioned in arguments, %compared to these topics semantic-relevant with the target, which is 
as shown in Figure~\ref{motivation}.

We use the global topic-word distribution
${\boldsymbol{T}} \in \mathbb{R}^{K \times V}$ to extract the explainable topics corresponding to each latent topic $k$,
i.e., a topic matrix where each $k$-th row ${\boldsymbol{T}}_{k,:} \in \mathbb{R}^V$ denotes a distribution over vocabulary words for the $k$-th topic.
As illustrated at the bottom of Figure \ref{fig:overall-structure}(b), some key terms relevant to the target are extracted. % to highlight the contribution of those topic words.

\paragraph{Topic Filtering}

Topic filtering returns $K$ lists of key terms, with each list corresponding to one of the latent topics, % for  each latent topic, 
i.e., ${\boldsymbol{l}} = [l_k|_{k=1:K}]$,
where $l_k$ consists of the top ${N}$ key terms for the $k$-th topic.
Firstly, the target words are filtered out with the mask $\boldsymbol{M}$,
i.e., $\boldsymbol{T} \odot \boldsymbol{M}$,
where $\odot$ denotes element-wise hadamard product,
$\boldsymbol{M} \in \mathbb{R}^{K \times V}$ is an indicator matrix where each column
$\boldsymbol{M}_{: ,i}\in\{1^K \quad {\textbf {if}}\quad \tau^{BOW}_i \neq 0; 0^K {\textbf{otherwise}}\}$ and
$\tau^{BOW}$ denotes the bag-of-words representation of the target $\tau$.
Secondly, the top ${N}$ key terms are selected for each topic:
\begin{align}
	{\boldsymbol{l}= \mbox{row}\_\arg\max[\boldsymbol{T} \odot \boldsymbol{M}]}_{1:N}
\end{align}
where ``row\_$\arg\max$'' is a function returning the indices of the top ${N}$ values from each row of the topic-word distribution.

\paragraph{Topic Extraction}

Given the $K$ lists of key terms ${\boldsymbol{l}} = [l_k|_{k=1:K}]$,
the most target-relevant topics $\boldsymbol{r}$ %with the top $N$ words 
are extracted by comparing the similarity between the target words and the topic words. %,which provides interpretability towards the target.
Specifically, let $\tau=\{\tau_{1},...,\tau_{N_{\tau}}\}$ be the target words and ${\boldsymbol{E}}$ be the word embedding matrix of a pre-training model (e.g., XLNet),
where $N_{\tau}$ is the number of the target words;

(I) the target representation is embedded as $\boldsymbol{e}_{\tau}={\rm emb\_lookup(\boldsymbol{E}}, \tau_{:}) \in \mathbb{R}^{N_{\tau} \times d}$
and $K$ lists of top ${N}$ key terms as $\boldsymbol{e}_{t_k}={\rm emb\_lookup(\boldsymbol{E}}, l_{k,:}) \in \mathbb{R}^{N \times d}$,
where the $d$ is the dimension of the word embedding.

(II) target-topic score is calculated separately for each word using cosine similarity,
$s_{\tau_i, l_k}= \frac{\sum {\rm row\_max}[\boldsymbol{e}_{\tau_i}\boldsymbol{e}_{l_k}^\top]_{1:N\times p}}{N_{\tau}}$,
where $0<p<1$ is the ratio that target should be similar to topic and $s_{\tau }\in \mathbb{R}^K$.

(III) the target-relevant topic words are extracted based on the target-topic score as $\boldsymbol{r}=[l_k\mid _{ k=\arg\max(s_{\tau})}]$.

\subsection{Argument Identification}
\label{argument identification}

To capture the semantic representation of the argument,
we employ a pre-trained XLNet ~\cite{DBLP:conf/nips/YangDYCSL19} model built on Transformer-XL \cite{dai-etal-2019-transformer} to encode the context information.
Specifically, given the concatenation of the argument sentence $\boldsymbol{x}^{\tau}$, the target $\tau$ and the extracted %explainable 
topics $\boldsymbol{r}$ as input,
the semantic representation $\boldsymbol{h}$ is generated by taking the $[CLS]$ representation of the last layer as the output.
Then the $\boldsymbol{h}$ is fed to the $MLP$ layer for argument classification $\hat{y}=\mbox{softmax}\big(MLP(\boldsymbol{h})\big)$.
At the same time, $\boldsymbol{h}$ is fed into the topic-argument mutual learning component to incorporate the sentence-level topic information.

\subsection{Topic-Argument Mutual Learning}
As illustrated in Figure~\ref{motivation},
we assume the \textit{support} or \textit{oppose} arguments are associated with more topics than \textit{none} argument,
which is reflected in the sentence-level latent topic distribution of the argument.
In this section,
we utilize mutual learning to incorporate sentence-level topic information into argument identification.
Following~\citep{gui2020multi, DBLP:conf/acl/SiZLSH20}, the latent topic distribution of the argument is leveraged to guide the learning of the argument representation.
In addition, the semantic representation information obtained from the language model could also be used to guide the generation of the latent topic distribution in the neural topic model with mutual learning.

Specifically,
given the sentence-level latent topic distribution $\boldsymbol{z}\in \mathbb{R}^K$ for the argument extracted from the neural topic model,
and the semantic information $\boldsymbol{h}$ obtained from the language model, for each argument, the similarity between $\boldsymbol{z}$ and $\boldsymbol{h}$ is calculated and maximized during the training.
Firstly, the semantic representation is affined into the latent topic vector $\boldsymbol{u}\in \mathbb{R}^K$ using the \textit{MLP} layer.
\begin{align}
	\boldsymbol{u}=\rm{softmax}\big(\emph{MLP}(\boldsymbol{h})\big)
\end{align}
Then, we use the similarity measurement metric $\mathcal{O}$, which is a harmonic mean of KL-divergence:
\begin{align}
	\mathcal{O}(\boldsymbol{u},\boldsymbol{z})=\sum \frac{1}{1+\frac{1}{\frac{1}{D_{K L}\left(\boldsymbol{u}\mid \boldsymbol{z}\right)}+ \frac{1}{D_{K L}\left(\boldsymbol{z}\mid \boldsymbol{u}\right)}}}
\end{align}

The mutual learning objective function based on the similarity function is defined as,
\begin{align}
	\mathcal{L}_m= \sum\limits_{D}\mathcal{O}(\boldsymbol{z}, \boldsymbol{u})
\end{align}

\subsection{Training Objective}
The neural topic model is trained by minimizing the VAE-loss with the mutual learning loss jointly.
\begin{align}
	\label{L_t}
	\mathcal{L}_t=\gamma \cdot \mathcal{L}_{m}+\mathcal{L}_{NTM}
\end{align}

The language model is trained by minimizing the cross-entropy loss and the mutual learning loss jointly,
\begin{align}
	\label{L_c}
	\mathcal{L}_c= \gamma \cdot \mathcal{L}_m + \sum_{D}\mathcal{L}_{cross} (y, \hat{y})%\limits_{(\boldsymbol{x}^{\tau},\tau,y ) \in D} \sum y \cdot log (\hat{y})
\end{align}
where $\boldsymbol{h}$ is fed into the $MLP$ layer for argument classification $\hat{y}=\mbox{softmax}\big(MLP(\boldsymbol{h})\big)$.

Overall, the training of the neural topic model and the language model is performed alternatively, which is shown in Algorithm~\ref{algorithm:1}.

\begin{algorithm}[t]
	\renewcommand{\algorithmicrequire}{\textbf{Input:}}
	\caption{Training Process}\label{algorithm:1}
	\begin{algorithmic}[1]
		\REQUIRE{arguments with targets and labels $\{\boldsymbol{x}^{\tau}, \tau, y\}$, pre-trained word embedding, the maximum training iterations $I$}
		\STATE{Initialize the model parameters;}
		\FOR{i=1 to $I$}
		\FOR{each batch of training instances}
		\STATE{Minimize the loss function in Eq.~\ref{L_t};}
		\ENDFOR
		\STATE{Obtain the latent topic distribution of the argument;}
		\STATE{Extract the explainable topics;}
		\FOR{each batch of training instances}
		\STATE{Minimize the loss function in Eq.~\ref{L_c};}
		\ENDFOR
		\ENDFOR
	\end{algorithmic}
\end{algorithm}

\section{Experimental Setting}
This section describes the datasets, baselines and implementation details in our experiments.

\paragraph{Datasets}
We evaluate the proposed TEAM on the public sentential argumentation mining collection (UKP ArgMin)\footnote{\url{https://tudatalib.ulb.tu-darmstadt.de/handle/tudatalib/2345}}~\citep{stab-etal-2018-cross},
which consists of 25,492 instances covering eight controversial targets
%(i.e.,~\textit{abortion, cloning, death penalty, gun control, marijuana legalization, minimum wage, nuclear energy, and school uniforms})
with number of 4,944, 6,195, 14,353 instances for~\textit{support, oppose} and \textit{none} arguments, respectively.
The statistics of the dataset are shown in Table~\ref{tab:dataset}.
The corpus provides a broad range of genres including news reports, editorials, blogs, debate forums, and encyclopedia articles.
The targets have been randomly selected from a list of controversial targets\footnote{\url{https://www.procon.org/}} \footnote{\url{https://www.questia.com/library/controversial-topics}}.
%The definition of argument in this dataset is a span of text with reasoning or evidence, which is able to either support or oppose a given target \citep{stab-etal-2018-cross}.

\paragraph{Baselines}
The following approaches are chosen as the baselines,
including the \textbf{LSTM-based models}, BiLSTM, Outer\_att and BicLSTM~\citep{stab-etal-2018-cross}, where the target semantic information is integrated into the model via concatenation, attention mechanism and recurrent unit, respectively;
\textbf{Lexicon-based models}, ClaimLex, SentimentLex, EmotionLex and WordNet~\citep{lin-etal-2019-lexicon}, which incorporate the lexicon information from different sources into the BiLSTM model; and
\textbf{BERT-based models}, BERT-base$_{target}$, BERT-large$_{target}$~\citep{DBLP:conf/acl/ReimersSBDSG19}, which incorporate the target semantic information into the language model.

\paragraph{Implementation Details}

The proposed model is evaluated on the in-target setting and the cross-target setting.
The in-target setting follows the conventional supervised paradigm while the cross-target setting aims to evaluate the robustness of the proposed model.
Therefore,
for the in-target setting, final evaluation results are the average over the ten-fold cross-valuation;
for the cross-target setting, training (70\%) and validation(10\%) data of seven targets are taken for training and parameter tuning respectively and the test data (20\%) of the eighth target is employed for testing,
the final evaluation results are the average results over the eight targets.
To evaluate the performance, the macro F1, precision and recall are employed as the evaluation metrics for both settings~\citep{DBLP:conf/acl/ReimersSBDSG19}.
During training, we use AdamW \cite{DBLP:journals/corr/abs-1711-05101} and Adam to optimize the parameters of the language model and the neural topic model with learning rates 2e-5 and 2e-3, respectively.
The mini-batch size is set to 16, $\gamma$ is set to 0.1, the vocabulary size used in NTM is 4,888, topic number $K$ is set to 10 and the pre-trained parameters for the language model are obtained from XLNet\footnote{\url{https://github.com/zihangdai/xlnet}}.

\begin{table}[]
	\resizebox{\columnwidth}{!}{%
		\begin{tabular}{l|c|ccc}
			\hline
			\textbf{Target}        & \textbf{Sentences} & \textbf{None} & \textbf{Support} & \textbf{Oppose} \\ \hline
			Abortion               & 3,929              & 2,427         & 680              & 822             \\
			Cloning                & 3,039              & 1,494         & 706              & 839             \\
			Death penalty          & 3,651              & 2,083         & 457              & 1,111           \\
			Gun control            & 3,341              & 1,889         & 787              & 665             \\
			Marijuana legalization & 2,475              & 1,262         & 587              & 626             \\
			Minimum wage           & 2,473              & 1,346         & 576              & 551             \\
			Nuclear energy         & 2,576              & 2,118         & 606              & 852             \\
			School uniforms        & 3,008              & 1,734         & 545              & 729             \\ \hline
			Total                  & 25,492             & 14,353        & 4,944            & 6,195           \\ \hline
		\end{tabular}}
	\caption{Statistics of the UKP ArgMin.}
	\label{tab:dataset}%
\end{table}

\section{Results}
\iffalse
	In this section,
	we evaluate the proposed TEAM from different aspects.
	Firstly, we compare the overall performance between  TEAM and the baselines in in-target setting and cross-target setting.
	Then we conduct the ablation study to show the effectiveness of using the explainable topics and the topic-argument mutual learning mechanism.
\fi

\begin{table}[t]\footnotesize
	\centering
	\begin{tabular}{l|ccccc}
		\hline
		\multicolumn{1}{c|}{\multirow{2}{*}{\textbf{Model}}} & \multicolumn{5}{c}{\textbf{UKP ArgMin}}                                                                                                                          \\ \cline{2-6}
		\multicolumn{1}{c|}{}                                & \textbf{F1}                             & \textbf{P+}               & \textbf{P-}               & \textbf{R+}               & \textbf{R-}                        \\ \hline
		BiLSTM                                               & .5337                                   & .4521                     & .4832                     & .2911                     & .4816                              \\
		BiCLSTM                                              & .5382                                   & .4185                     & .4469                     & .3860                     & .4813                              \\ \hline
		ClaimLex                                             & .5684                                   & .4736                     & .5075                     & .3756                     & .5011                              \\
		SentimentLex                                         & .5718                                   & .4937                     & .5125                     & .3590                     & .5240                              \\
		EmotionLex                                           & .5695                                   & .4920                     & .5036                     & .3524                     & .5264                              \\
		WordNet                                              & .5788                                   & .4846                     & .5191                     & .3724                     & .5235                              \\ \hline
		BERT-base                                            & .6710                                   & .5742                     & .5937                     & .5811                     & .6250                              \\
		BERT-large                                           & .7008                                   & .5954                     & .6426                     & .6568                     & .6637                              \\ \hline \hline
		\textit{TEAM}                                        & \textbf{.7243}                          & \textbf{.6177}            & \textbf{.6814}            & \textbf{.6820}            & .6723                              \\
		\textit{-ML}                                         & \multicolumn{1}{l}{.7113}               & \multicolumn{1}{l}{.6123} & \multicolumn{1}{l}{.6474} & \multicolumn{1}{l}{.6425} & \multicolumn{1}{l}{\textbf{.7018}} \\
		\textit{-ET}                                         & \multicolumn{1}{l}{.7135}               & \multicolumn{1}{l}{.6116} & \multicolumn{1}{l}{.6722} & \multicolumn{1}{l}{.6725} & \multicolumn{1}{l}{.6691}          \\
		\textit{-ET \& ML}                                   & \multicolumn{1}{l}{.6815}               & \multicolumn{1}{l}{.5813} & \multicolumn{1}{l}{.5976} & \multicolumn{1}{l}{.6137} & \multicolumn{1}{l}{.6354}          \\\hline
	\end{tabular}
	\caption{Results of each model using in-target evaluation on the test sets.
		\textbf{BERT-base/large}: BERT-base/large$_{target}$,
		\textbf{F1}: macro F1, \textbf{P}: precision, \textbf{R}: recall, \textbf{+}: support argument, \textbf{-}: oppose argument,
		\textbf{ML}: mutual learning,
		\textbf{ET}: explainable topic information.
		Bold values indicate the best result.}
	\label{tab:in-target}%
\end{table}

% \usepackage{multirow}

\begin{table}[t]\footnotesize
	\centering
	\begin{tabular}{l|ccccc}
		\hline
		\multicolumn{1}{c|}{\multirow{2}{*}{\textbf{Model}}} & \multicolumn{5}{c}{\textbf{UKP ArgMin}}                                                                                                                          \\ \cline{2-6}
		\multicolumn{1}{c|}{}                                & \textbf{F1}                             & \textbf{P+}               & \textbf{P-}                        & \textbf{R+}               & \textbf{R-}               \\ \hline
		BiLSTM                                               & .3796                                   & .3484                     & .4710                              & .0963                     & .2181                     \\
		Outer-att                                            & .3873                                   & .3651                     & .4696                              & .1042                     & .2381                     \\
		BicLSTM                                              & .4242                                   & .2675                     & .3887                              & .2817                     & .4028                     \\
		BicLSTM$_{E}$                                        & .3924                                   & .2372                     & .4381                              & .0317                     & .3955                     \\
		BicLSTM$_{B}$                                        & .4243                                   & .3431                     & .4397                              & .1060                     & .4275                     \\ \hline
		BERT-base                                            & .6128                                   & .5048                     & .5313                              & .4698                     & .5795                     \\
		BERT-large                                           & .6325                                   & .5535                     & .5843                              & .5051                     & .5594                     \\ \hline \hline
		\textit{TEAM}                                        & \textbf{.6748}                          & \textbf{.6199}            & .6063                              & \textbf{.5296}            & \textbf{.6340}            \\
		\textit{-ML}                                         & \multicolumn{1}{l}{.6654}               & \multicolumn{1}{l}{.6136} & \multicolumn{1}{l}{.6249}          & \multicolumn{1}{l}{.5152} & \multicolumn{1}{l}{.6135} \\
		\textit{-ET}                                         & \multicolumn{1}{l}{.6665}               & \multicolumn{1}{l}{.6156} & \multicolumn{1}{l}{\textbf{.6286}} & \multicolumn{1}{l}{.5150} & \multicolumn{1}{l}{.5881} \\
		\textit{-ET \& ML}                                   & \multicolumn{1}{l}{.6214}               & \multicolumn{1}{l}{.5173} & \multicolumn{1}{l}{.5421}          & \multicolumn{1}{l}{.4737} & \multicolumn{1}{l}{.5683} \\\hline
	\end{tabular}
	\caption{Results of each model using cross-target evaluation on the test sets.
		\textbf{BicLSTM$_{E/B}$}: BicLSTM$_{ELMo/BERT}$.
		Bold values indicate the best result.}
	\label{tab:cross-target}%
\end{table}

\subsection{Overall Performance}

\paragraph{In-target Setting}
Table~\ref{tab:in-target} reports the overall performance of our model against baselines.
As shown in Table~\ref{tab:in-target},
among the baselines,
BERT-large$_{target}$ gives the best results.
Our model significantly outperforms all the baselines in all evaluation metrics.
It is worth pointing out %pointed 
that our model outperforms BERT-large$_{target}$ by over 2\% in macro F1. This is a significant improvement considering that BERT-large is a more powerful model in representation learning compared to XLNet which is used in our proposed approach. % while  representation ability of the BERT-large is stronger than XLNet-base which our model used. 
It demonstrates the effectiveness of introducing the topic information into argument representation learning.

The lower half of Table~\ref{tab:in-target} presents the ablation study by removing various components from TEAM.
It can be observed that by removing different components,
the performance of TEAM drops consistently in different metrics except for the recall of the \textit{oppose} arguments without mutual learning, in which recall is improved. %the topic diversity information. It might be attributed to that the extracted explainable topics provide the additional information however weaken the ability of capturing the topic diversity information.
One possible reason is that, in the in-target setting, there are common topics both in \textit{oppose} arguments and \textit{support} arguments,
mutual learning might confuse the information contained in these arguments.

\paragraph{Cross-target Setting}

To evaluate the robustness of our model,
we conduct cross-target experiments to evaluate how well the model is generalized to an unknown target,
which reflects a real-life argument search scenario.
The results under the cross-target setting are shown in Table~\ref{tab:cross-target}.
Overall, the BERT-based models have higher performance than LSTM-based models because of the rich contextualized information. Our TEAM scores the highest in terms of macro F1 ( 67.48\%), which is 4.23\% absolute points higher than the BERT-large$_{target}$ baseline.
It can be observed that the performance of the proposed TEAM under cross-target setting drops nearly %close 
5\% compared to that under the in-target setting,
which shows the difficulty of AM under the cross-target setting.
%In addition, 
We also notice that TEAM has higher improvement against the strong baseline BERT-large$_{target}$ (i.e., 4.23\% vs. 2.06\%) compared to that under the in-target setting. %It can be attributed to 
One possible reason is that the topic information %, i.e., the explainable topic information and the topic diversity information, are 
is more crucial for AM under the cross-target setting. %   more target-specific representation for better capturing the information across the targets. I NOT QUITE UNDERSTAND.

The lower half of Table~\ref{tab:cross-target} presents the results of the ablation experiments.
With the removal of different components in our model,
the performance dropped nearly 1\% in terms of macro F1 compared with the full version.
However, the precision of \textit{oppose} arguments increases by %yields nearly 
2\% without the topic information. %drop.
The reason might be that in cross-target setting,
there are diverse ways to express the opposition (e.g., \textit{mock})
and the topic information might introduce noise, making it more difficult for the model to identify the \textit{oppose} arguments for unseen targets.

\begin{table}[]\small
	\centering
	\begin{tabular}{c|ccccc}
		\hline
		\multirow{2}{*}{\textbf{Topics}} & \multicolumn{5}{c}{\textbf{UKP ArgMin}}                                                                                                                                         \\ \cline{2-6}
		                                 & \multicolumn{1}{l}{\textbf{F1}}         & \multicolumn{1}{l}{\textbf{P+}} & \multicolumn{1}{l}{\textbf{P-}} & \multicolumn{1}{l}{\textbf{R+}} & \multicolumn{1}{l}{\textbf{R-}} \\ \hline
		10                               & .7243                                   & .6177                           & .6814                           & .6820                           & .6723                           \\
		20                               & .7233                                   & .6151                           & .6845                           & .6798                           & .6741                           \\
		30                               & .7149                                   & .6219                           & .6506                           & .6546                           & .7000                           \\
		40                               & .7125                                   & .6051                           & .6447                           & .6754                           & .6982                           \\
		50                               & .7160                                   & .6473                           & .6456                           & .6239                           & .7009                           \\ \hline
	\end{tabular}
	\caption{Evaluation of TEAM with different topic numbers.}
	\label{tab:topic-number}%
\end{table}

\subsection{Effect of Topic Number}
Table~\ref{tab:topic-number} shows the performance of our model with the different number of topics ranging between 10-50 on the UKP ArgMin.
As shown in Table~\ref{tab:topic-number}, the performance of the proposed approach TEAM decreases slightly with the increase of the topic number. This is expected since the total number of targets in our data is 8. Nevertheless, the model with the different number of topics %given the ground truth 8, which  
still outperforms the baselines.
We choose the model with the topic number $K=10$,
which has the highest macro F1 score.

\begin{table}[]\small
	\centering

	\begin{tabular}{lccc}
		\toprule
		%\multicolumn{1}{m}{\multirow{2}[3]{*}{\textbf{Models}}} & \multicolumn{3}{c}{\textbf{UKP}}                                                        \\
		\cmidrule{2-4}     & \multicolumn{1}{l}{$C_p$} & \multicolumn{1}{l}{$C_v$} & \multicolumn{1}{l}{NPMI} \\
		\midrule
		LDA                & 0.009                     & 0.363                     & 0.019                    \\
		NVDM               & 0.084                     & 0.416                     & -0.014                   \\
		NTM-R              & 0.280                     & 0.408                     & 0.059                    \\
		% GSM          & 0. & 0. & 0. &       &       &       &       &       &  \\
		ProdLDA            & 0.267                     & 0.433                     & 0.053                    \\
		\midrule
		% TKIM-A & 0.296 & 0.460  & 0.060  \\
		\textit{our TEAM}  & \textbf{0.331}            & \textbf{0.462}            & \textbf{0.066}           \\
		\textit{-ML}       & 0.296                     & 0.460                     & 0.060                    \\
		\textit{-ET}       & 0.270                     & 0.461                     & 0.054                    \\
		\textit{-ET \& ML} & 0.260                     & 0.430                     & 0.049                    \\
		\bottomrule
	\end{tabular}%
	\caption{Topic coherence score comparison of our proposed model and baselines. }
	\label{tab:topic-model}%
\end{table}%

\begin{table}[t]\footnotesize
	\centering

	\begin{tabular}[c]{c| p{0.55\columnwidth}l}

		\toprule
		\textbf{Target}          & \multicolumn{1}{c}
		{\textbf{Top 10 Words from TEAM}}                                                                                          \\
		\midrule
		{Gun control}            & weapon, firearm, ownership, citizen, assault, law, violence, restriction, conceal, defense      \\ \hline
		{Abortion}               & woman, mother, baby, fetus, womb, hood, life, soul, right, bear                                 \\ \hline
		{School unifroms}        & student, dress, teacher, discipline, administrator, test, policy,academic, wear, standard       \\ \hline
		{Minimum wage}           & worker, employment, increase, pay, job, poverty, economy, price, business, labor                \\ \hline
		{Cloning}                & cell, clone, human, stem, unethical, dignity, animal, reproduction, gene, organ                 \\ \hline
		{Nuclear energy}         & plant, power, reactor, radioactive, renewable, fuel, electricity, fossil, emission, safe        \\ \hline
		{Death penalty}          & woman, punishment, crime, murder, execute, drug, rate, family, action, work                     \\ \hline
		{Marijuana legalization} & drug, recreational, illicit, criminal, medical, alcohol, colorado, claliform, regulate, revenue \\

		\bottomrule
	\end{tabular}%
	\caption{Top 10 words for 8 target extracted from the UKP ArgMin using TEAM model.}
	\label{tab:topic-word}%
\end{table}%
\subsection{Analysis of Topic Information}
We analyze the topic information generated from the NTM from two perspectives:
1) qualitative analysis of the explainable topics extracted from the model;
2) quantitative analysis of the topic coherence score.

\begin{figure*}[t]
	\centering
	\includegraphics[scale=0.56]{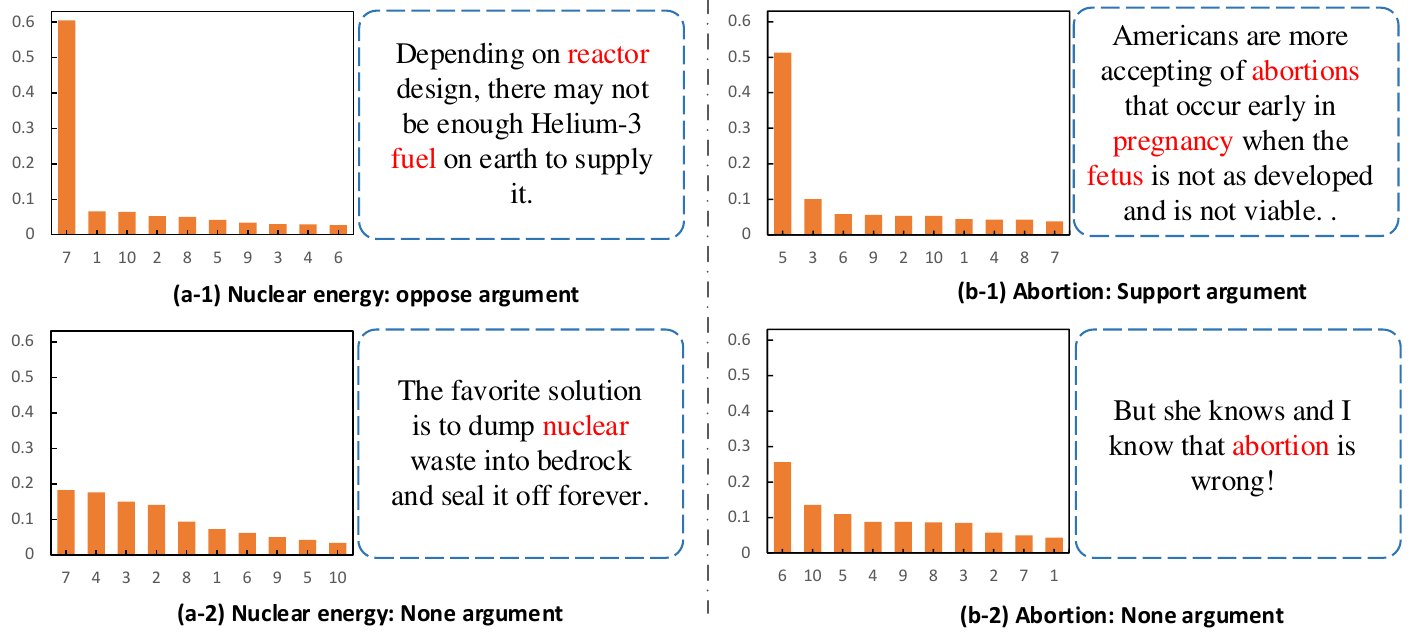}
	\caption{The sentence-level latent topic distribution and argument sentences with explainable topics (\textcolor{red}{red}) extracted by our TEAM.}
	\label{fig:topic-diversity}
\end{figure*}
\paragraph{Explainable Topics}
%To obtain an intuitive impression on the explainable topics,
For qualitative evaluation,
we present the top 10 topic words for 8 targets extracted by our model in Table~\ref{tab:topic-word}.
It can be observed that the topics extracted by TEAM are coherent and interpretable towards the target (e.g., \textit{ownership, law} for \textit{gun control}, and \textit{discipline, policy} for \textit{school uniforms}),
which can be considered as the complementary information for the target.

\paragraph{Topic Coherence}
To evaluate the quality of the topics,
following~\citet{DBLP:conf/aistats/WangGWSHPSC18},
we employ the topic coherence (NPMI scores, $C_v$, $C_p$ measures) of top 5/10/15/20 topic words for each topic as  evaluation metrics (the higher is better).
Specifically,
the average coherence score among topics is calculated to compare our method with the following baselines:
LDA~\citep{DBLP:journals/jmlr/BleiNJ03}, NTM-R~\citep{ding-etal-2018-coherence}, NVDM~\citep{DBLP:journals/corr/MiaoYB15} and ProdLDA~\citep{DBLP:conf/iclr/SrivastavaS17}.

\iffalse
	\begin{table*}[]
		\renewcommand\arraystretch{1.3}
		%\renewcommand\tabcolsep{3.0pt}
		\centering
		\begin{tabular}{l|l|c|c|c}

			\hline
			\textbf{Target} & \textbf{Examples}                                                                                                                            & $y$ & $\hat{y}$ & $\bar{y}$ \\ \hline \specialrule{0em}{1pt}{3pt}
			School uniforms & \parbox[l]{9cm}{When \textcolor{red}{dressed} neatly and seriously, \textcolor{red}{students} tend to behave seriously.}                     & $y$ & $\hat{y}$ & $\bar{y}$ \\ \hline \specialrule{0em}{1pt}{3pt}
			Nuclear energy  & \parbox[l]{9cm}{Depending on \textcolor{red}{reactor} design, there may not be enough Helium-3 fuel on \textcolor{red}{earth} to supply it.} & $y$ & $\hat{y}$ & $\bar{y}$ \\ \hline \specialrule{0em}{1pt}{3pt}
			Minimum wage    & \parbox[i]{9cm}{And \textcolor{red}{workers} have a bit more cash in their \textcolor{red}{paychecks} besides.}                              & $y$ & $\hat{y}$ & $\bar{y}$ \\ \hline \specialrule{0em}{1pt}{3pt}
		\end{tabular}
		\caption{hahahahhaa}
		\label{fig:case study}
	\end{table*}
\fi

Table~\ref{tab:topic-model} presents the results of our model against the baselines. It can be observed that
1) the proposed TEAM gives the best overall results on all evaluation metrics compared to other baselines, which demonstrates the benefit of incorporating the semantic representations of arguments into the topic model learning;
2) TEAM without mutual learning ($-ML$) leads to a slight decrease of the coherence scores, showing that the semantic information learned by the language model can indeed help generating more coherence topics.
%incorporating the explainable topic information and the sentence-level topic information does improve the ability of the model to capture the topic information in the argument.

\paragraph{Visualization}
To provide an intuition of the topics extracted by our TEAM,
we further visualize the latent topic distribution and the explainable topics of the argument in four cases in Figure~\ref{fig:topic-diversity},
in which TEAM  predicted correctly while BERT-large$_{target}$ failed.
It can be observed that compared to \textit{none} arguments,
\textit{support} or \textit{oppose} arguments contain more target-related topics and exhibit a peaked %perform a sharper 
topic distribution,
which verifies our hypothesis that target-related topic information is indeed important for the identification of reasoned arguments. %the motivation in our paper.

\subsection{Error Analysis}
We randomly select 100 incorrectly predicted instances in both the in-target setting and the cross-target setting and summarize the main errors.
The first type of error %caused by our model 
is that our model tends to predict the %compared with 
\textit{oppose} arguments wrongly compared to \textit{support} arguments %,  are predicted incorrectly with a higher probability 
in both settings. One possible reason is that \emph{oppose} opinions are expressed in a more diverse ways compared to \emph{support} arguments, which makes the identification of \emph{oppose} arguments more difficult. %It might attribute to the various ways of expressing their oppose attitude, which needs more accurate detection approach.
The second type of error is caused by implicit and ambiguous %that 
\textit{opinions} expressed in arguments.
Even humans would have problems in identifying this category of arguments correctly.
In addition, certain arguments need to be identified by incorporating external knowledge, such as the background of a celebrity. This is difficult for models relying on text only. %, which is hard to be identified only based on the texts.

\section{Conclusion}
We have presented an explainable Topic-Enhanced Argument Mining model that incorporates the topic information for information-seeking AM.
The explainable topics extracted from the topic-word distribution are used to augment the target information.
Moreover,
we propose topic-argument mutual learning to incorporate the latent topic distribution into the language model for capturing the sentence-level topic information.
%In addition,
The argument semantic representation is used to guide the neural topic model for generating diverse %distinguishable 
topics with mutual learning.
Empirical results validate the effectiveness of our model in topic extraction and argument identification.
In future work, %it is also important to 
we will explore ways to combine the external knowledge and the topic information for more accurate argument mining.

\bibliography{anthology,custom}
\bibliographystyle{acl_natbib}

\appendix

\end{document}